# DeepDiveAI: Identifying AI Related Documents in Large Scale Literature Dataset

**Zhou Xiaochen[1,2] Liang Xingzhou[1] Zou Hui[3] Lu Yi[1,4] Qu Jingjing[1,*]**

[1]Shanghai Artificial Intelligence laboratory [2]The University of Hong Kong
[3]Shanghai University [4]King's College London

Abstract: In this paper, we propose a method to automatically classify AI-related documents from large-scale literature databases, leading to the creation of an AI-related literature dataset, named DeepDiveAI. The dataset construction approach integrates expert knowledge with the capabilities of advanced models, structured across two global stages. In the first stage, expert-curated classification datasets are used to train an LSTM model, which classifies coarse AI related records from large-scale datasets. In the second stage, we use Qwen2.5 Plus to annotate a random 10% of the coarse AI-related records, which are then used to train a BERT binary classifier. This step further refines the coarse AI related record set to obtain the final DeepDiveAI dataset. Evaluation results demonstrate that the entire workflow can efficiently and accurately identify AI-related literature from large-scale datasets.

## 1. Introduction

The development of artificial intelligence (AI) is having a transformative impact on society, but the research of the science of AI is just beginning. In the summer of 1956, John McCarthy, Marvin Minsky, Claude Shannon, Nathaniel Rochester McCarthy and other scientists held a meeting at Dartmouth College in the United States to discuss how to simulate human Intelligence with machines, and first proposed the concept of "Artificial Intelligence (AI)"[1]. Today, the rapid development of AI is bringing subversive changes to a wide range of fields. Especially in the rapid development of large language models, the capability emergence and generalization performance of large models have brought new development and application prospects of artificial intelligence,

---

✉ Corresponding author. Contact qujingjing@pjlab.org.cn.

Funded by Science and Technology Innovation 2030 - 'New Generation Artificial Intelligence' Major Project.

enabling scientific research[2], manufacturing[3], education[4], healthcare,[5] and public administration[6]. The importance of this phenomenon is also reflected in that, with the continuous emergence of the application value of artificial intelligence, more and more organizations and researchers begin to try to explore the potential value[7]. However, despite AI's rise as a global area of focus, our understanding of its innovation model and developmental principles remains significantly underdeveloped.

Looking back at the development of artificial intelligence, it has experienced two distinct "AI winters"[8], [9]. The first occurred from the late 1960s to the 1980s, when AI research faced significant funding cuts due to limitations in computing power and the failure to meet ambitious targets. The second winter, in the mid-1990s, was partly caused by hardware manufacturers' inability to keep up with the specialized demands of expert systems [10]. A common feature of these AI winters is that, despite the high levels of attention and resources initially directed toward AI, the field eventually cooled off after a period of heightened interest. This phenomenon reflects our insufficient understanding of the patterns and sustainable trajectories of AI innovation. A key manifestation of this challenge is the lack of high-quality public datasets in AI research, which hinders systematic and strategic analysis of AI innovation activities, as it lacks the necessary foundational data.

For social scientists, policymakers, and industry stakeholders, analyzing AI scientific research texts is crucial for understanding the current state of AI research, its evolution, and its future trajectory. Text is an important data source for social computing[11]. By mining textual data, we can extract deeper insights into implicit topics, relational networks, and other latent structures within the data[12]. For example, latent topic structures can be reappeared by Latent Dirichlet Allocation (LDA)[13], [14]. Moreover, research on the integration of AI with other disciplines requires large-scale AI-related scientific data as a foundation. Shi and Evans have proposed methods for assessing the content and contextual novelty of scientific literature by analyzing textual features, such as keywords and citation networks[15]. This approach offers an effective tool for understanding AI's innovation trajectory and interdisciplinary development, but it requires comprehensive textual support. All these in-depth social computing analyses rely on high-quality, large-scale datasets derived from the AI literature. In this context, our research goal is to construct such a dataset, providing a solid foundation for advancing scientific research in artificial intelligence.

## 2. Literature review

"Artificial Intelligence" is a broad term often used to describe the simulation of human intelligence. The term "artificial intelligence" was first coined in 1955 by John McCarthy, who defined it as "the science and engineering of making intelligent machines"[16], [17]. OECD defined the AI system as a machine-based system that can, for a given set of human-defined objectives, make predictions, recommendations or decisions influencing real or virtual environments[18]. In terms of goals, AI seeks to enable machines to replicate cognitive functions associated with the human mind, allowing them to perform complex tasks that typically require human intervention. Following Agrawal et al.'s characterization of AI as a "predictive machine," Miric et al. define it as a collection of statistical predictive techniques[19], [20]. However, many foundational papers in AI do not explicitly use the term "artificial intelligence" itself; instead, they focus on specific technologies such as "machine learning"[21], [22]. As an umbrella term, "artificial intelligence" may represent a vast iceberg, with much of the underlying technology and field still hidden beneath the surface.

The identification of AI related literature has evolved from traditional literature taxonomy. Historically, literature classification have primarily relied on journal-level categorization, which is based on scale and expert knowledge. However, recent advancements in research have shifted the granularity of classification from the journal level to the article level, exploring pathways toward automated classification methods. Traditionally, bibliometric methods defined the subject of the literature based on journal-level categorization[23]. This approach necessitates expert knowledge to assign attribute markers, and with the increasing volume of journal publications, it has led to a substantial number of misclassifications[24], [25]. Nonetheless, journal-level classification remains advantageous in that it offers a cost-effective method for broad categorization, which can be useful for preparing training data for more precise, article-level classifiers. In recent years, however, the focus has shifted toward article-level classification, reflecting a refinement in the granularity of the classification process. For instance, the pre-publication platform arXiv mandates that authors specify the classification of their articles upon submission, with artificial intelligence being one of its key subfields[26]. However, given that AI is an inherently cross-disciplinary field, relying solely on papers labeled "AI" may overlook its integration with related domains and real-world scenarios, which are crucial drivers of AI's sustainable development. Consequently, identifying "AI-related" literature based exclusively on classification labels is insufficient and fails to capture the full breadth of the field.

In parallel, classification methods are transitioning from manual to automated approaches. While manual classification benefits from expert intervention, which enhances accuracy, it is limited by inefficiencies in processing large-scale datasets and incurs high costs. In the realm of AI literature, automated classification methods are increasingly recognized for their growing importance. One key issue, however, lies in the difficulty of definitively determining whether a paper belongs to the AI field [27]. Furthermore, the interdisciplinary nature of AI introduces ambiguity and inconsistency in applying classification labels[28]. Essentially, the complexity of scientific inquiry makes it challenging to establish a singular, comprehensive, or flawless classification structure[29].

This situation has led researchers to predominantly classify AI-related studies using data-driven methods, rather than rule-based methods[30]. The limitations of rule-based methods include the need for manual rule design and feature extraction, which entails significant engineering effort. Furthermore, these methods may struggle to adapt to the dynamic evolution of both rules and academic knowledge. In contrast, the data-driven approach offers the advantage of enabling meaningful automated classification[31]. As a result, recent research has focused primarily on exploring automated classification methods based on text data. Among these methods, Long Short-Term Memory (LSTM) networks have been identified as an important technique[32], [33], [34].

Sachini et al. (2022) utilized a BERT-based model to classify AI papers in the Scopus database, achieving an accuracy of 96.5%, however the data size of their actual classification was relatively small[23]. Similarly, Miric et al. (2023) compared the performance of various unsupervised machine learning models for AI patent classification, finding that BERT outperformed LSTM. Nevertheless, their study was based on a smaller dataset with only 3,200 training samples, which limits the model's classification accuracy on large-scale data[20]. With the growing capabilities of large language models (LLMs) in understanding language, there has been increasing interest in their use for text classification tasks. Toney-Wails et al. (2024) employed fine-tuned LLMs (GPT) for automated AI domain labeling but highlighted concerns regarding high computational costs and reliability issues[35]. Wang et al. (2023) found that LLMs can perform classification tasks effectively without the need for extensive task-specific training, using zero-shot text classification[36].

Existing studies have demonstrated that LSTM, BERT, and large language models (LLMs) are effective methods for classifying AI-related literature. However, each of these methods has distinct advantages and limitations. LSTM is known for their relatively fast training and inference speeds, making them computationally efficient[37]. However, its ability to capture complex linguistic nuances is not as good as advanced models such as BERT [38]. While BERT offers significantly higher accuracy, it requires substantial computational resources, particularly during large-scale applications[39], [40]. On the other hand, large language models, such as GPT, provide excellent accuracy and context understanding, but their use is associated with high computational costs and slower inference times[41].

For large-scale datasets, the differences in model capabilities highlight the need to balance computational efficiency, scalability, and reliability when developing automated classification methods. This paper aims to leverage the strengths of various models to construct datasets efficiently and reliably for AI-related scientific literature.

## 3. Methods

This paper proposes a hierarchical and refined classification methodology by leveraging LSTM,

BERT, large language models (LLMs), and expert knowledge. A detailed overview is provided in FIG. 1. The process unfolds in several key stages:

1. Data Preparation: We began by extracting global research paper data from the Intelligent Innovation Dataset (IIDS), a hyperscale database that integrates six interrelated datasets spanning nearly 120 years[42]. This comprehensive resource includes information on research papers, citation relationships, patent details, patent legal statuses, and funding data, serving as the foundational source for subsequent analysis.

2. Simple Classification Set Construction: we solicit expert knowledge to construct a simple classification set. This dataset includes AI-related papers from several widely recognized and reputable AI conferences, classified as "AI-related," and a random sample from IIDS, categorized as "non-relevant."

3. Coarse Classification with LSTM: The LSTM model is trained using the simple classification set. Leveraging LSTM's fast training and inference capabilities, we perform a rough classification on the large dataset, generating a preliminary set of "AI-related" records.

4. LLM Annotation of Coarse Sampling Set: We randomly select 10% of the samples from the initial rough classification set and apply annotation using a Large Language Model (LLM), generating a refined LLM-labeled subset for further analysis.

5. Refined Classification with BERT: The LLM-labeled set is used to train a BERT binary classifier. By leveraging BERT's superior natural language understanding, we reclassify the rough classification set, yielding the final, refined database, which we call it DeepDiveAI dataset.

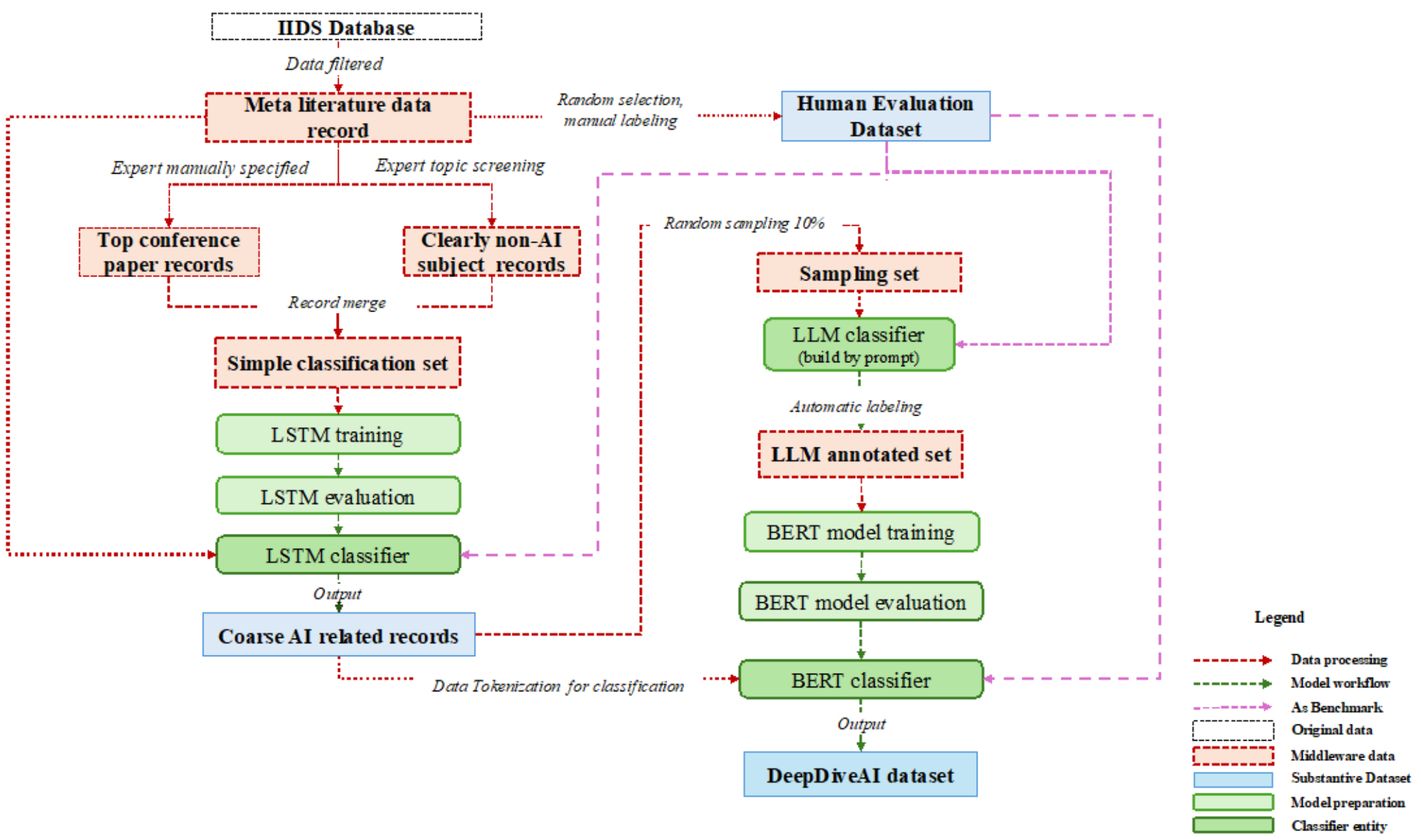

**Figure1. Workflow of AI-Related Documents Classification**

To ensure the reliability and validity of the entire classification framework, we construct an Expert Evaluation Dataset (EED) consisting of 5,000 samples, based on expert knowledge, to assess the performance of LSTM, LLM, and BERT. The results demonstrate that our proposed methodology effectively constructs an accurate and reliable classifier capable of determining whether scientific research papers belong to the AI domain, based on expert judgment.

### 3.1 Data Preparation

The original scientific research literature data used in this paper is sourced from the IIDS database, which provides extensive coverage of data related to scientific achievements and patents. The database currently includes over 92.31 million records across various types of academic outputs, such as global academic papers, scientific research projects, patents, and other forms of scholarly contributions. These records span 27 major disciplines and 334 sub-disciplines, encompassing natural sciences, medical sciences, social sciences, and life sciences.

**Table 1** provides a description of the key features of the dataset:

| Features | Description |
|---|---|
| id | A unique identifier for each paper |
| Source Title | The title of the journal or conference proceedings |
| Title of the paper | The title of the paper |
| Paper abstract | The abstract or summary of the paper |
| Author Name | The name of the paper's contributor |
| Author ID | A unique identifier for the author |
| Author Keywords (JSON) | A list of keywords provided by the author, in JSON format |
| Publication Year | The year in which the paper was published |

**Table 1. Training dataset description**

For this study, we extract records of academic papers from the IIDS database as the source data. We focus specifically on two key fields: paper_title and paper_summary. Whether employing expert manual judgment or automated methods, the relevance of a paper to the field of AI is determined based on the combined text information from these two fields.

### 3.2 Simple Classification Set Construction

In constructing the simple classification dataset, expert knowledge was leveraged to determine the scope of AI-related papers. To ensure scientific rigor, we invited seven experts with extensive experience in AI research to help formulate the classification rules. The process consisted of the following steps:

(1) Preliminary screening of candidate conferences and journals. Initially, we compiled a list of potentially relevant conferences and journals from the IIDS database based on citation rates and industry evaluations. This served as the starting point for the classification process.

(2) Expert Feedback. We then gathered input from the experts through interviews. The key

questions posed included:

- Which conferences are most representative of the field of artificial intelligence?
- Are there additional relevant journals that should be included?
- How can we scientifically define the selection criteria for "non-relevant" samples? Drawing on their research experience and knowledge of current trends, the experts provided detailed responses to these questions.

(3) Refinement of expert consensus. Based on the feedback from the experts, we distilled the commonalities to define the criteria for identifying "AI-related" sources. This helped to establish a shared understanding of what constitutes an AI-related paper. As a result, three leading conferences were selected as sources for "AI-related" papers: the Conference on Neural Information Processing Systems (NeurIPS), the International Conference on Learning Representations (ICLR), and the International Conference on Machine Learning (ICML). Furthermore, experts suggested broadening the scope when selecting "non-relevant" samples, advocating for random sampling across diverse disciplinary backgrounds to avoid a skewed distribution.

**Conference on Neural Information Processing Systems (NeurIPS)**: NeurIPS, founded in 1987, is one of the leading international conferences in machine learning and computational neuroscience. Held annually every December, it gathers top researchers to present cutting-edge work in deep learning, reinforcement learning, generative models, and more.

**International Conference on Representation of Learning (ICLR)**: ICLR was founded in 2013 and is the premier conference focused on deep learning and representation learning. Spearheaded by Yoshua Bengio and Yann LeCun, one of the big Three in deep learning, ICLR quickly became a cutting-edge platform for exploring neural networks, unsupervised learning, and new machine learning architectures.

**International Conference on Machine Learning (ICML)**: ICML was founded in 1980 and is the flagship conference in the field of machine learning. ICML covers a wide range of topics from theoretical research to practical applications, including active learning, clustering, online learning, reinforcement learning, supervised learning, deep learning, optimization, probabilistic reasoning, and more.

To ensure the quality of the dataset and meet the training requirements for the LSTM model, the final training dataset consisted of 44,000 papers, evenly divided into 22,000 AI-related and 22,000 non-AI papers. The AI-related papers were sourced from the three leading AI conferences, while the non-AI papers were randomly selected from a diverse range of disciplines, including agriculture, biological sciences, immunology, microbiology, arts and humanities, health professions, nursing, and dentistry. This balanced and varied selection aimed to provide a robust foundation for model training.

### 3.3 Coarse Classification with LSTM

To perform the preliminary classification of a large literature dataset, we employed a deep learning model based on a Long Short-Term Memory (LSTM) network for rough classification. The aim of this approach is to quickly identify potentially AI-related papers within a large-scale dataset, thereby significantly reducing the computational burden in the subsequent refinement stage. The process unfolds as follows:

#### 3.3.1 Data preprocessing and vectorization

Before training the model, we preprocess and vectorize the simple classification set data. The dataset is split into a training set and a test set in an 80:20 ratio, with the training set used to optimize the model parameters and the test set employed to evaluate the model's generalization ability. For text vectorization, we utilize the **CountVectorizer** method, which transforms the text into a term frequency matrix. This method builds a vocabulary, assigning a unique index to each token, and converts the text into a matrix where each dimension represents the frequency of a specific token in the corresponding document, providing a structured representation of the text.[43], [44]

#### 3.3.2 LSTM model architecture design

The LSTM model effectively captures contextual semantic relationships in text through its unique memory unit mechanism. The architecture includes an input layer that converts vectorized text into a time-step sequence for LSTM processing. The LSTM layer, with 128 hidden units, models long-term dependencies by selectively retaining relevant information. Batch normalization follows to improve stability and convergence speed. The extracted features are passed through a fully connected layer for binary classification, with the output layer using a sigmoid activation function. To prevent overfitting, a dropout layer with a 0.5 ratio is applied for regularization. The ReLU activation function is used to introduce nonlinearity, improving the model's ability to learn complex patterns.

#### 3.3.3 LSTM training configuration and inference

The model was trained using the Adam optimizer, configured with a learning rate of 0.001 and a weight decay of 5e-4. Training spanned 50 epochs, with a batch size of 32 chosen to balance robust learning iterations and computational efficiency. The model's performance is evaluated on the test set after each epoch, with key metrics such as accuracy, precision, recall, and F1-score computed to assess its effectiveness. During evaluation, the loss and accuracy on the test data were calculated, ensuring an unbiased assessment of the model's generalization capability. The results of the accuracy and loss are presented in the following figure:

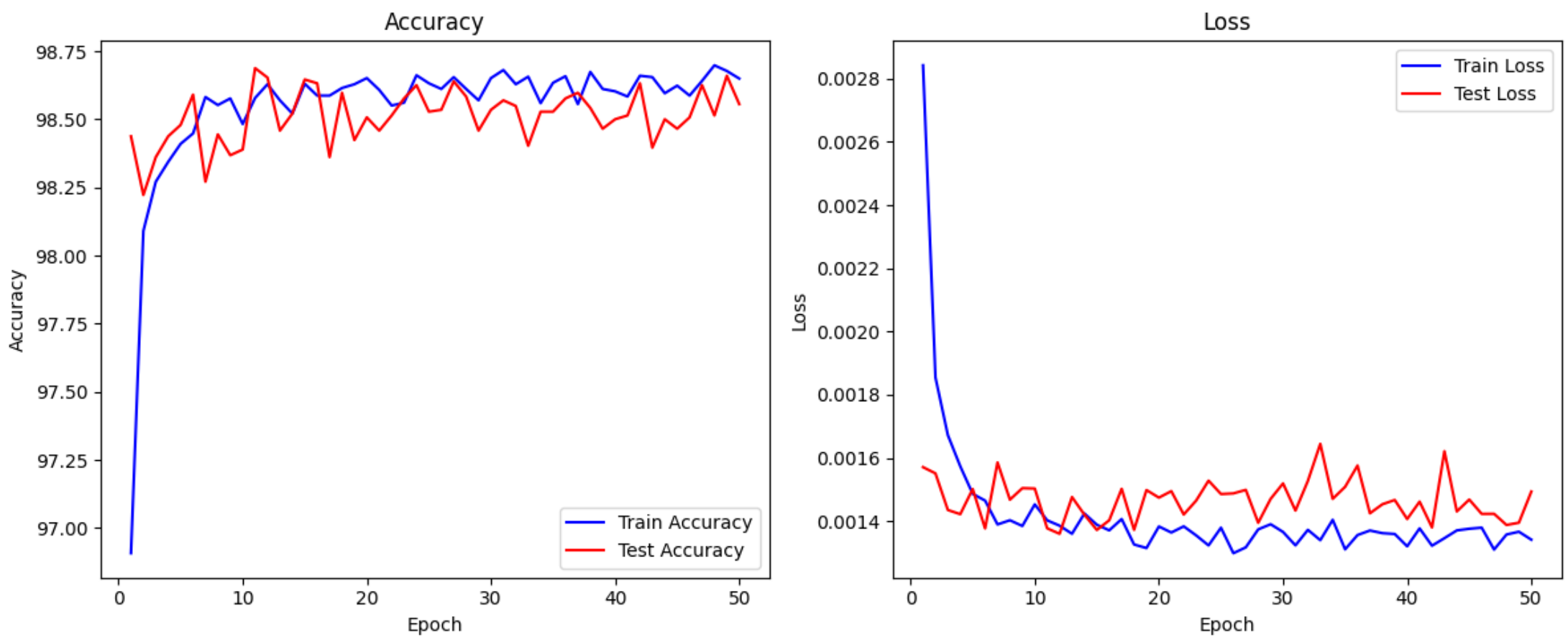

**Figure2. Accuracy and Loss results of LSTM Model**

The results indicate that the training accuracy increases steadily, reaching about 98.75%, and the test accuracy stabilizes around 98.6%. The training loss decreases sharply in the initial stages before leveling off, with the test loss remaining relatively stable but exhibiting minor fluctuations. The trend demonstrates LSTM exhibits high classification accuracy on the test set. By applying LSMT to the data set, we obtained 9479,846 rough classification data.

**3.4 LLM Annotation of Coarse Sampling Set**

We randomly selected 10% of the data from the LSTM Coarse Classification set, resulting in a subset of 947,985 entries. After filtering out samples with missing titles or abstracts, we utilized Tongyi Qianwen 2.5 for automatic labeling. Developed by Alibaba Cloud, Qwen-2.5 is part of the Qwen (Tongyi Qianwen) series of large language models, which have been pre-trained on vast datasets and exhibit strong performance across various benchmarks[45]. For this task, we specifically employed the qwen2.5-plus variant. The classification process involved concatenating prompt keywords with the paragraphs to be evaluated, including both the title and abstract, and submitting these concatenated prompts to the LLM in a single dialogue round. The model was then instructed to produce the annotation results formatted as JSON.

The prompt words used are as follows:

You are an expert in the field of artificial intelligence. Please judge whether the following text belongs to scientific or technological research in the field of artificial intelligence. The explanation of the reasons needs to be concise. Please output in the following format: If yes, please output

"{"status":1,"reason":"your reason for judgment"}",

If not, please output

"{"status":0,"reason":"your reason for judgment"}".

For ambiguous content, please set status to 0.5. The following is the material you need to judge:

Qwen-2.5-Plus API was used for annotation generation, and 879537 valid annotation data were generated. Its distribution is as follows:

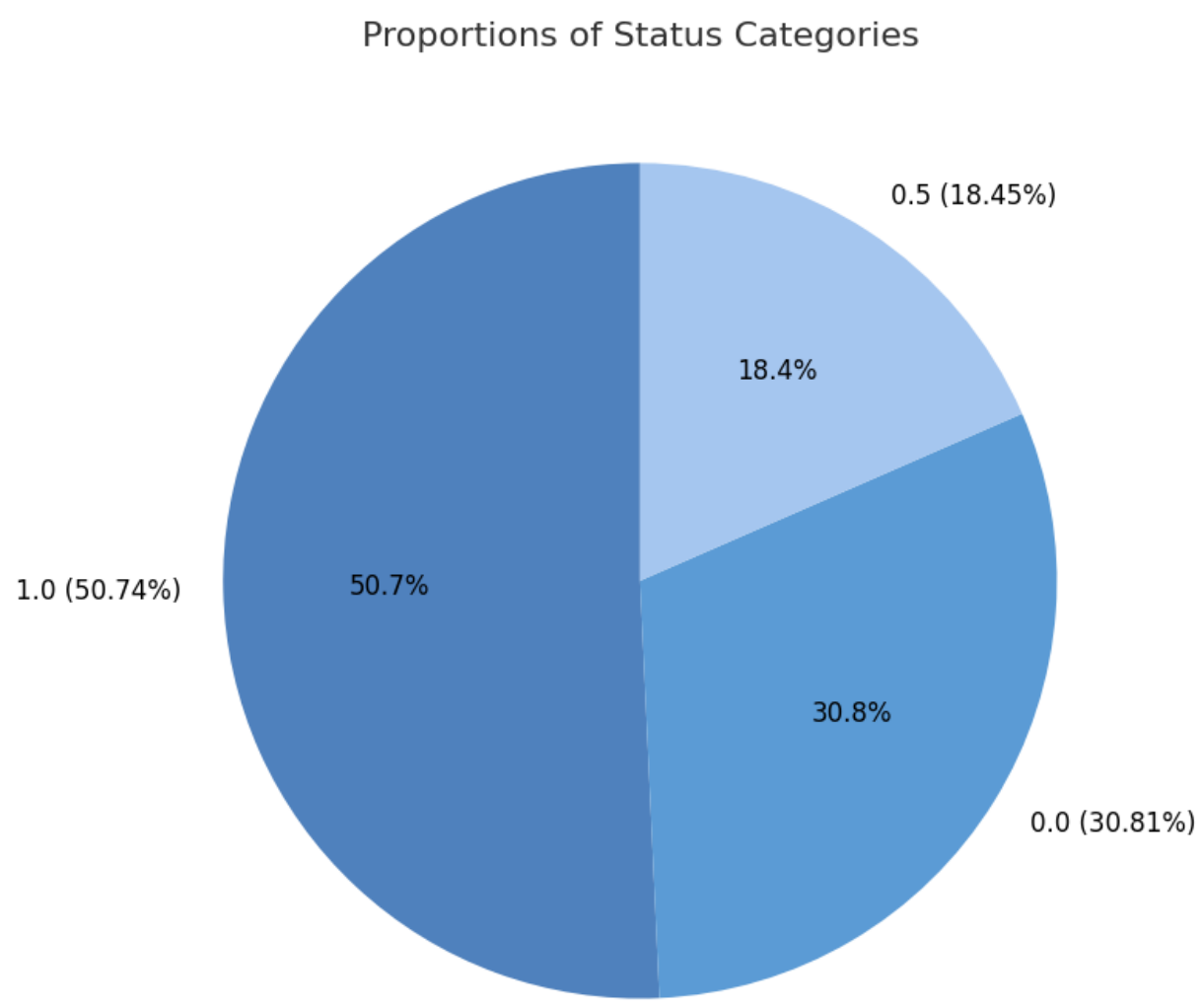

**Figure3. Qwen2.5-plus Annotation Results Distribution**

### 3.5 Refined Classification with BERT

Based on the large-scale labeled dataset generated by the LLM, the BERT model is further trained for fine-grained classification to enhance classification accuracy. BERT (Bidirectional Encoder Representations from Transformers) is a pre-trained transformer-based language model that captures the context of words in text by considering both the left and right surrounding words simultaneously[46]. In this study, we use the pre-trained model **bert-base-uncased**, developed by Google, and fine-tune it to suit the classification task. To rigorously determine whether a sample belongs to the AI domain, we treat samples marked with a value of 0.5 by the LLM as 0. The key hyperparameters are configured as follows:

- Number of training cycles: The model was trained for 15 cycles (epochs) to fully learn the data features.
- Batch size: The batch size was 64 for both training and evaluation to take into account model performance and memory usage.
- Learning rate: The initial learning rate is set to 2e-5 and a linear decay strategy is used during training.
- Warm-up steps: Warm-up the learning rate within the first 5000 steps (about 10% of the total) to ensure stability during the training process.
- Weight decay: A weight decay coefficient of 0.01 is used to prevent overfitting of the model.
- Gradient Accumulation: Set the number of gradient accumulation steps to 2 to simulate a

larger effective batch size, thereby improving training efficiency.

- Early stop strategy: The early stop mechanism of patience=5 is configured to avoid the model to continue training when the performance is stagnant, thus saving computing resources.

In the actual training, the model was stopped early after 7 rounds of training. The loss curves of Bert model training set and verification set are shown in the figure:

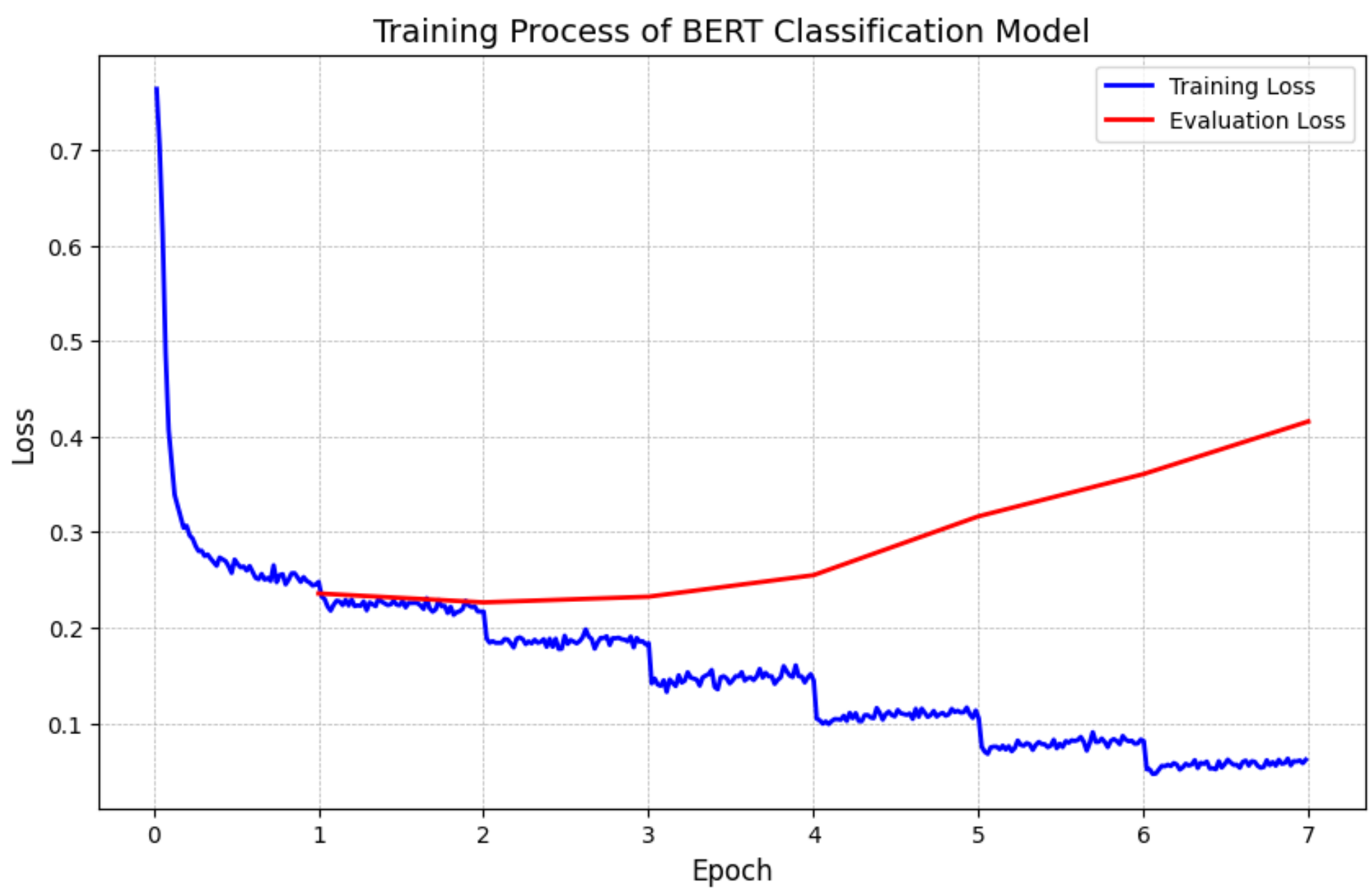

**Figure4. Loss of BERT Model**

After the training is completed, the BERT model is used for tokenization and inference of the whole rough classification set to form the final AI-related database.

### 3.6 Technical Verification

To assess the reliability of the classification model and the consistency with expert judgment, we constructed an Expert Evaluation Dataset (EED), which was randomly sampled from CCF-recommended journals for classification evaluation by domain experts. The performance of LSTM, LLM, and BERT models was then evaluated.

The EED dataset consists of 1,963 randomly selected articles from CCF journals. CCF journals are high-impact academic journals and conference proceedings in computer science and related fields, as recommended by the China Computer Federation[47]. While these journals contain research relevant to artificial intelligence, it is not immediately clear whether each article belongs specifically to the field of AI. This characteristic of the EED dataset makes it well-suited for evaluating how closely a classification model aligns with human decision-making in complex classification tasks.

For dataset construction, we first extracted 2,000 article records from the IIDS database that corresponded to journals included in the CCF list before 2023. To ensure data completeness, we removed samples with missing abstracts, leaving a final count of 1,963 articles. Subsequently, experts in the field of artificial intelligence manually verified the classification of these articles, resulting in the final EED.

Using EED as a benchmark, we validated LSTM, QWEN, and BERT using the classification_report tool in sklearn.metrics. Below are the summarized findings:

| Category | precision | recall | f1-score | accuracy |
|---|---|---|---|---|
| 0 | 0.53 | 0.43 | 0.47 | 0.70 |
| 1 | 0.75 | 0.82 | 0.79 | |

**Table. 2 LSTM classification report**

Despite showing some proficiency in category 1, LSTM's lower performance in category 0 suggests it may be better suited for coarse-grained classification tasks rather than capturing complex patterns.

| Category | precision | recall | f1-score | accuracy |
|---|---|---|---|---|
| 0 | 0.84 | 0.79 | 0.82 | 0.89 |
| 1 | 0.91 | 0.93 | 0.92 | |

**Table. 3 Qwen2.5-plus classification report**

The Qwen2.5-plus model showcased superior performance across both categories, effectively identifying positive class samples while minimizing the mislabeling of negative classes as positives. This demonstrates the potential of large-scale models in addressing imbalanced datasets.

| Category | precision | recall | f1-score | accuracy |
|---|---|---|---|---|
| 0 | 0.97 | 0.37 | 0.54 | 0.80 |
| 1 | 0.77 | 0.99 | 0.87 | |

**Table. 4 Bert classification report**

In this experiment, the BERT model achieved a high recall rate of 0.99 for Category 1, almost perfectly identifying samples within this category. This high recall aligns closely with the task's objectives, which aim to maximize the identification of AI-related papers. Furthermore, BERT demonstrated high precision in Category 0 (precision of 0.97), showcasing its capability to effectively minimize false positives. Our final BERT model meets the dual requirements of high recall and low false positives, making it a reliable tool for automatic classification and well-suited to the task's objectives.

**4. Results and Findings**

By further sifting through the coarse classification set using BERT, we ended up with 4,426,296 AI-related papers, spanning from 1956 to 2023. The reason for using 2023 as the cut-off point is that the 2024 publication data has not yet been fully aggregated, and 1956 was chosen as the starting point because it was at the Dartmouth conference that the concept of "artificial intelligence" was first formally proposed, marking a new era of scientific exploration and technological progress.

The DeepDiveAI dataset contains the underlying information for the papers, including data such as the title and abstract of the papers, as well as keywords provided by the authors. The specific database details are listed in Table 4.

| Features | Data type | Description |
|---|---|---|
| id | int64 | Sequential identifier ranked from 1 to 9479868 |
| paper_title | string | The title of the paper |
| source_title | string | The title of the journal or conference proceedings |
| paper_summary | string | A summary or abstract of the paper |
| paper_authors_name | string | Names of authors who contributed to the paper |
| paper_authors_id | string | Unique ID for identifying authors |
| pub_year | int64 | Paper publication year |
| cited_num | int64 | Number of times the paper has been cited |
| author_keyword_json | string | A JSON-formatted list of keywords provided by the authors |
| doi | string | Unique identifier that provides a link for online access |
| paper_type | string | Type of papers |
| language | string | Language of publication |
| affilations_info | string | Detailed affiliation information |

**Table 5. DeepDiveAI database description**

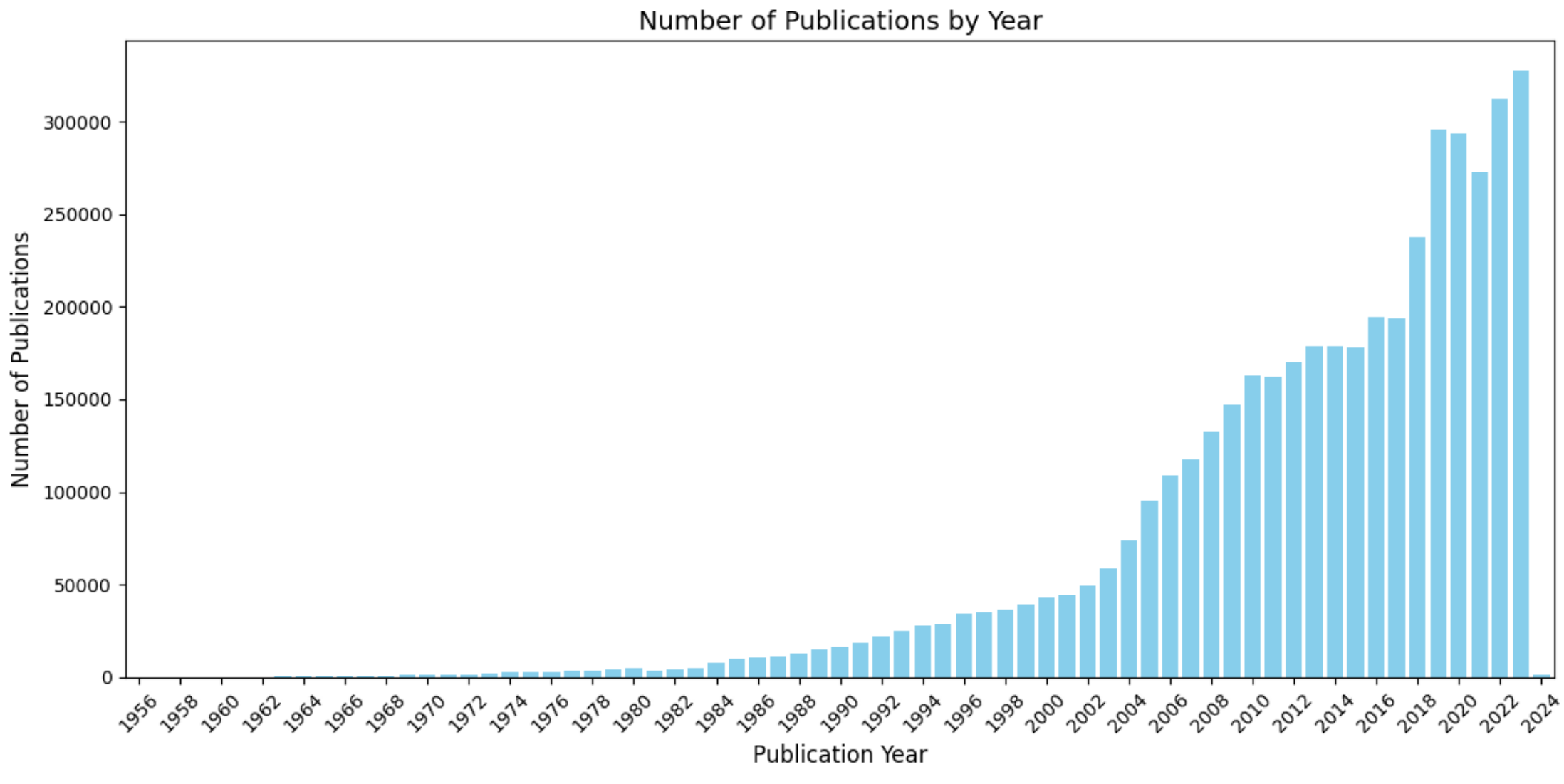

**Figure 5. AI documents published per year.**

The graph shows the number of AI-related documents published each year from 1956 to 2023. There is a clear upward trend, with a significant increase in publications in the 2000s, peaking around 2020. This trend highlights the rapid expansion of AI research and its growing importance in various fields.

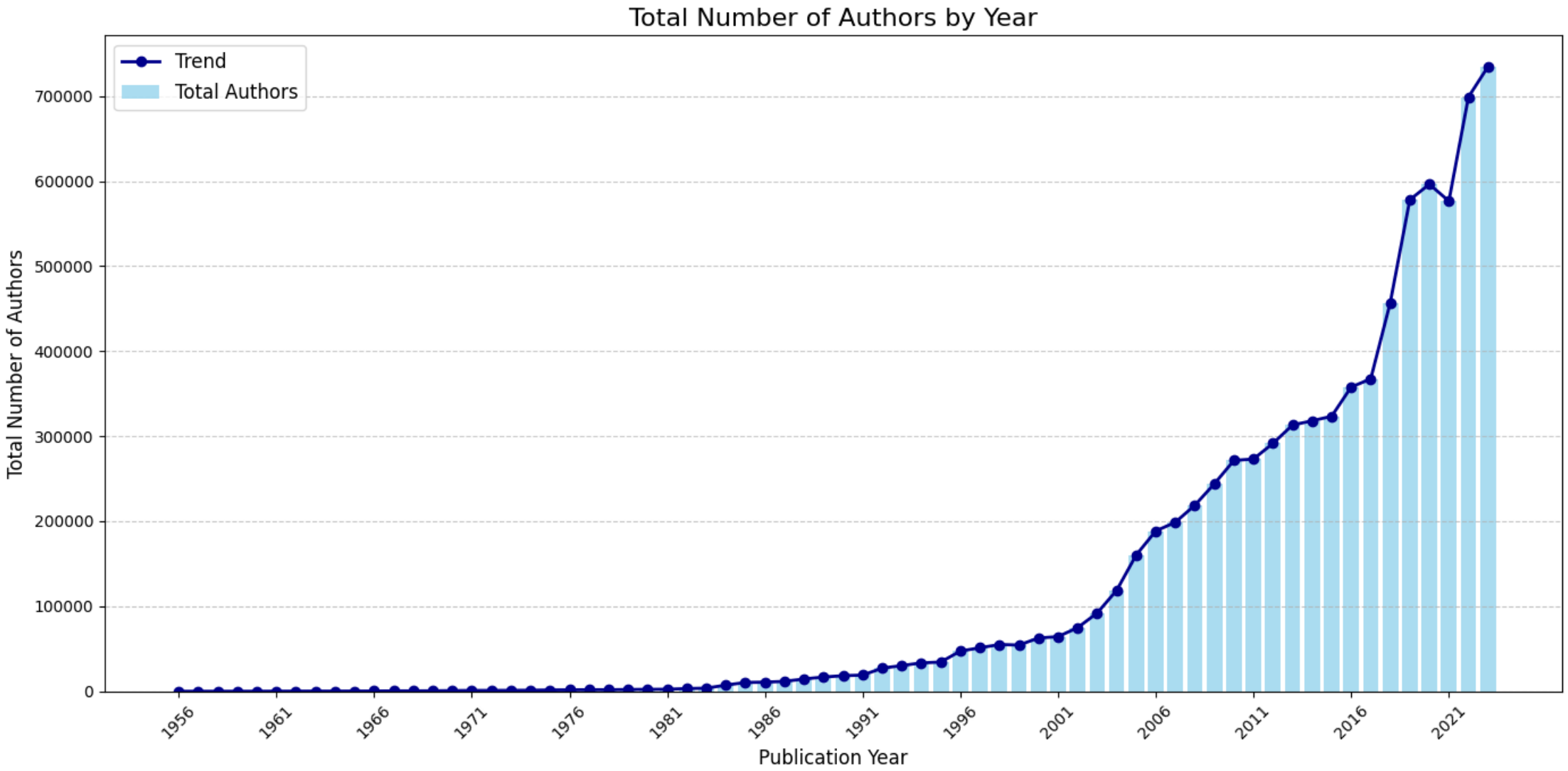

**Figure 6. Number of authors per year.**

The graph shows the number of authors contributing to AI-related publications each year from 1956 to 2023. A significant increase in the number of authors is observed starting around 2005. The growth becomes particularly steep from 2010 onwards, reflecting the expanding interest and involvement in AI research globally. The peak in the recent years further underscores the broadening participation in AI-related studies across various disciplines.

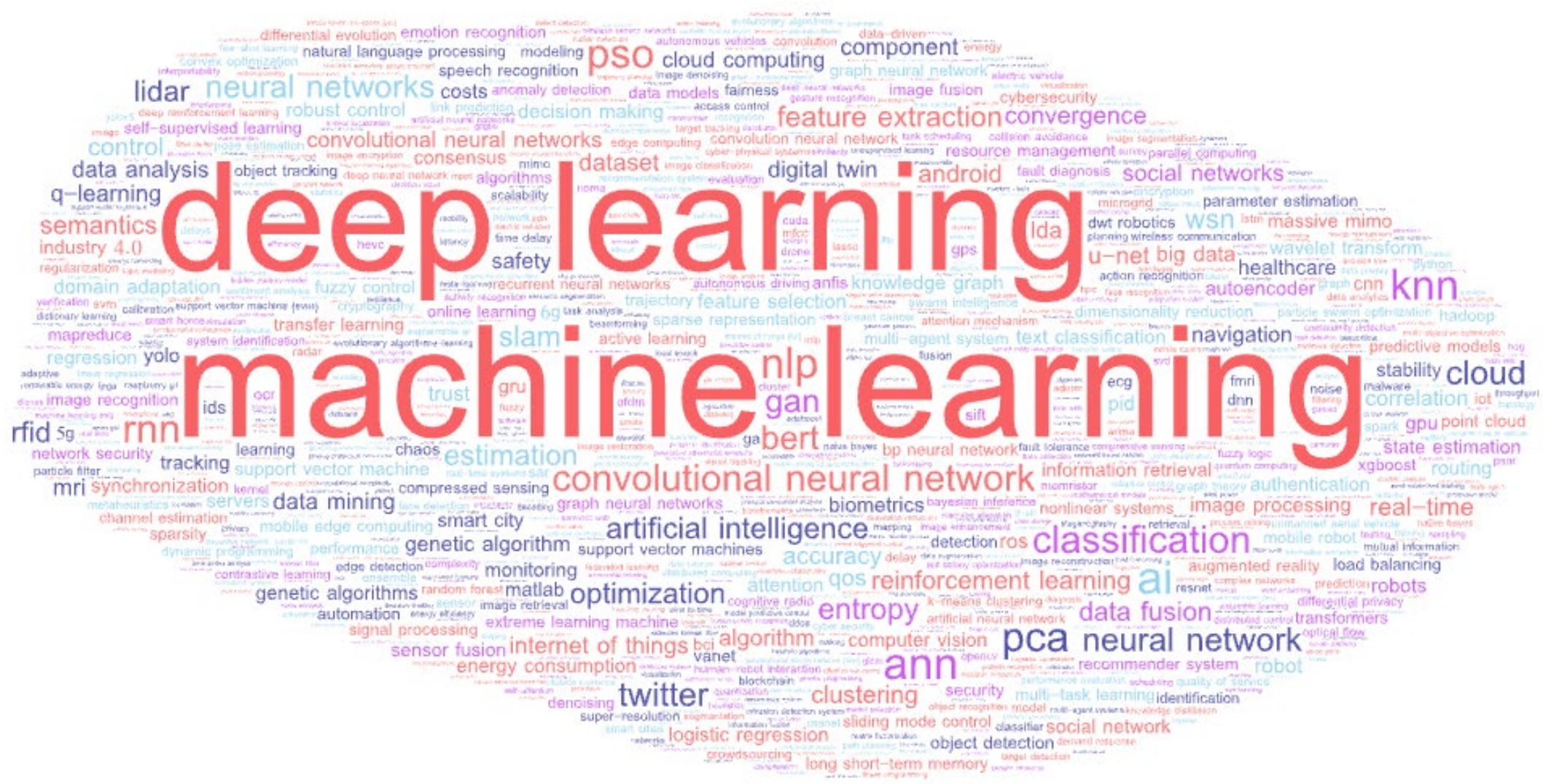

**Figure 7. Word cloud of keyword from 2016 to 2023 (top 1500).**

Prominent terms include machine learning (109311 occurrences), deep learning (107095 occurrences), convolutional neural network (26480 occurrences), and neural network (23278 occurrences), reflecting significant trends and research focuses in AI and related fields during this period. These keywords emphasize the importance of advanced learning algorithms and their applications in various domains.

## 5. Conclusion

The primary goal of this study is to develop and validate the DeepDiveAI database, a comprehensive and credible resource of AI-related literature. By leveraging the comparative advantages of models such as LSTM, LLM, and BERT, we designed an automated classification method for AI-related literature, progressing from coarse-grained classification to fine-grained classification. This approach allowed us to identify AI-related datasets consistent with expert judgment in the IIDS database, providing a valuable resource for understanding the trends and development of AI research.

Future work will focus on using the DeepDiveAI database to uncover deeper research patterns and trends, further enhancing our understanding of the evolution of AI research. Additionally, exploring advanced neural network architectures and integration techniques will help improve classification performance and offer more nuanced insights into the field. Ensuring the continued reliability and comprehensiveness of the DeepDiveAI database remains a top priority, and we aim to maintain it as a crucial resource for the AI research community.

Acknowledgment

This work was supported by Science and Technology Innovation 2030 - 'New Generat ion Artificial Intelligence' Major Project.